\documentclass[letterpaper, 10 pt, conference]{ieeeconf}
\IEEEoverridecommandlockouts
\usepackage{cite}
\usepackage{amsmath,amssymb,amsfonts}
\usepackage{algorithmic}
\usepackage{graphicx}
\usepackage{tabularx}
\usepackage{subcaption}
\usepackage{textcomp}
\usepackage{xcolor}
\def\BibTeX{{\rm B\kern-.05em{\sc i\kern-.025em b}\kern-.08em
    T\kern-.1667em\lower.7ex\hbox{E}\kern-.125emX}}
\begin{document}

% Extras
\newtheorem{nullhyp}{Null Hypothesis}

\title{\LARGE \bf A Blended Human-Robot Shared Control Framework to Handle Drift and Latency\\
\thanks{*This research is supported by the Department of Energy under Award Number DE-EM0004482, by the National Aeronautics and Space Administration under Grant No. NNX16AC48A issued through the Science and Technology Mission Directorate and by the National Science Foundation under Award Nos. 1451427, 1544895, 1649729.}
\thanks{Authors are affiliated with the Department of Electrical \& Computer Engineering, Northeastern University, Boston, MA, 02215
{\tt\small$^{1}$abouallaban.a@husky.neu.edu, \{$^{2}$v.dimitrov, $^{3}$t.padir\}@northeastern.edu}}
\thanks{Example video can be found at https://tinyurl.com/y7rue2k7}
}

\author{Anas Abou Allaban$^{1}$, Velin Dimitrov$^{2}$, Ta\c{s}k{\i}n~Pad{\i}r$^{3}$}

\maketitle

\begin{abstract}
Maximizing the utility of human-robot teams in disaster response and search and rescue (SAR) missions remains to be a challenging problem. This is due to the dynamic, uncertain nature of the environment and the variability in cognitive performance of the human operators. By having an autonomous agent share control with the operator, we can achieve near-optimal performance by augmenting the operator's input and compensate for the factors resulting in degraded performance. What this solution does not consider though is the human input latency and errors caused by potential hardware failures that can occur during task completion when operating in disaster response and SAR scenarios. In this paper, we propose the use of blended shared control (BSC) architecture to address these issues and investigate the architecture's performance in constrained, dynamic environments with a differential drive robot that has input latency and erroneous odometry feedback. We conduct a validation study (n=12) for our control architecture and then a user study (n=14) in 2 different environments that are unknown to both the human operator and the autonomous agent. The results demonstrate that the BSC architecture can prevent collisions and enhance operator performance without the need of a complete transfer of control between the human operator and autonomous agent. 
\end{abstract}

% -------------------- %
% --- Introduction --- % 
% -------------------- %
\section{Introduction}
Shared control within the context of robotics has been studied as early as the late 1980s \cite{sheridan:92, hayati:89, anderson:96, lacey:00, crandall:02, wasson:03, simpson:97}. Early work was based on implementations where the robot control was transferred between teleoperation and autonomous modes with the goals of addressing transmission time-delays \cite{hayati:89}, developing context-aware navigation procedures, and characterizing efficiency in human-robot interaction \cite{crandall:02}.

Now, the research in shared control seeks to address the question: how can we treat human-robot teams as a whole? The collaborative architecture of shared control makes it ideal for applications where the human's control is augmented by the autonomous agent or vice-versa. As such, shared control is a topic of interest in assistive and service robotics \cite{gopinath17, Broad2017LearningDynamics, Yu2003AnElderly}. 

In disaster response and SAR applications, human teleoperation of a robot is usually characterized by high latency, poor visual \& sensory feedback, and limited time for task completion. Furthermore, the human operator must avoid collisions and account for dynamic changes in the environment. The challenges poised against the human operator lead to degraded performance and situational awareness \cite{Yanco2004quotWherePlatform}. On the other hand, autonomous agents alone have difficulty traversing these dynamic environments. Rather than having the agent and operator work separately, we can have human-robot teams collaborate via a shared control architecture on a certain task, complementing one other's inefficiencies.

\begin{figure}[!t]
    \centerline{\includegraphics[width=\columnwidth]{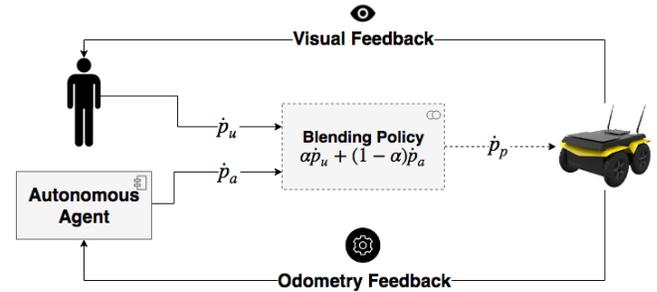}}
    \caption{The system architecture and input flow to the operator and autonomous agent. Note that odometry and visual feedback is erroneous in the second study.}
    \label{fig:sc_arch}
\end{figure}

There has, however, been limited research on the application of model-based shared control in the context of robotics navigating in a constrained environment such as a factory or disaster environment, specifically when environmental disturbances such as latency (i.e. poor signal) and erroneous sensor feedback (i.e. damaged hardware) are not taken in consideration. Approaches towards semi-autonomy using reinforcement learning \cite{Kartoun2010AAlgorithm} and hierarchical reinforcement learning \cite{Doroodgar2010TheRobots, Doroodgar2010AEnvironments} were made, however, with a focus on deciding \textit{when} the autonomous agent should cede control to a human operator and not necessarily augmenting their input in real time. Only recently has \cite{Reddy2018SharedLearning} conducted an analysis on blending human and autonomous agent inputs using model-free deep reinforcement learning. \cite{Reddy2018SharedLearning} also consider an autonomous agent with delayed input, but do not consider erroneous feedback from the robot or delayed input from the operator, variables that we account for in our study.

While a machine learning approach could be taken to design a shared control architecture that addresses latency and noise, it is difficult to obtain enough training data and construct high fidelity simulators to collect said data in order to have a high performance learning algorithm \cite{Reddy2018SharedLearning}. Furthermore, it is also difficult to incorporate all physics-based failure modes that can result in the drift, a focal point in this research.  Thus, a model-based shared control architecture would be more fit to the task.

The lack of understanding of how shared control approaches degrade under latency and drift disturbances is a significant barrier for their adoption. The main contribution of this work is quantifying the degradation of performance with input latency and drift disturbances in combination with a linear blended shared control approach. 

The paper is organized as follows. In section \ref{sec:approach} we discuss the shared control architecture used and the overall methodology for validating the architecture in various environments. In section \ref{sec:implementation}, we formalize our blended shared control architecture and our choice of the blending parameter. \ref{sec:setup} details the experimental setup of the user study to quantify the performance of the approach in time delay and drift scenarios. Section \ref{sec:validation_results} and \ref{sec:dynamic_results} analyzes the performance of the approach in two real world tasks: a doorway traversal and navigation in a constrained environment, respectively. Finally, Section \ref{sec:futurework} highlights the future work necessary before shared control could be reliably used in real-world scenarios involving time delay and drift.
% --------------------------------------------------------------------------------

% ------------------- %
% -- Methodology & -- %
% ---- Approach ----- %
% ------------------- %
\section{Approach \& Methodology}
\label{sec:approach}

The collaborative nature of blended shared control makes it a control architecture suitable for when the operator is impaired in some manner. If the operator's input would lead to a task failure or damage to the robot, the autonomous agent can ``intervene'' and prevent a failure or collision without totally relinquishing the operator's control. By allowing for a continuous transition back to the operator, the human-robot team can resume task completion with minimal downtime. This is critical in the context of a disaster response and SAR mission where the time to complete the task is essential.

\begin{figure}[htbp]
    \centerline{\includegraphics[width=\columnwidth]{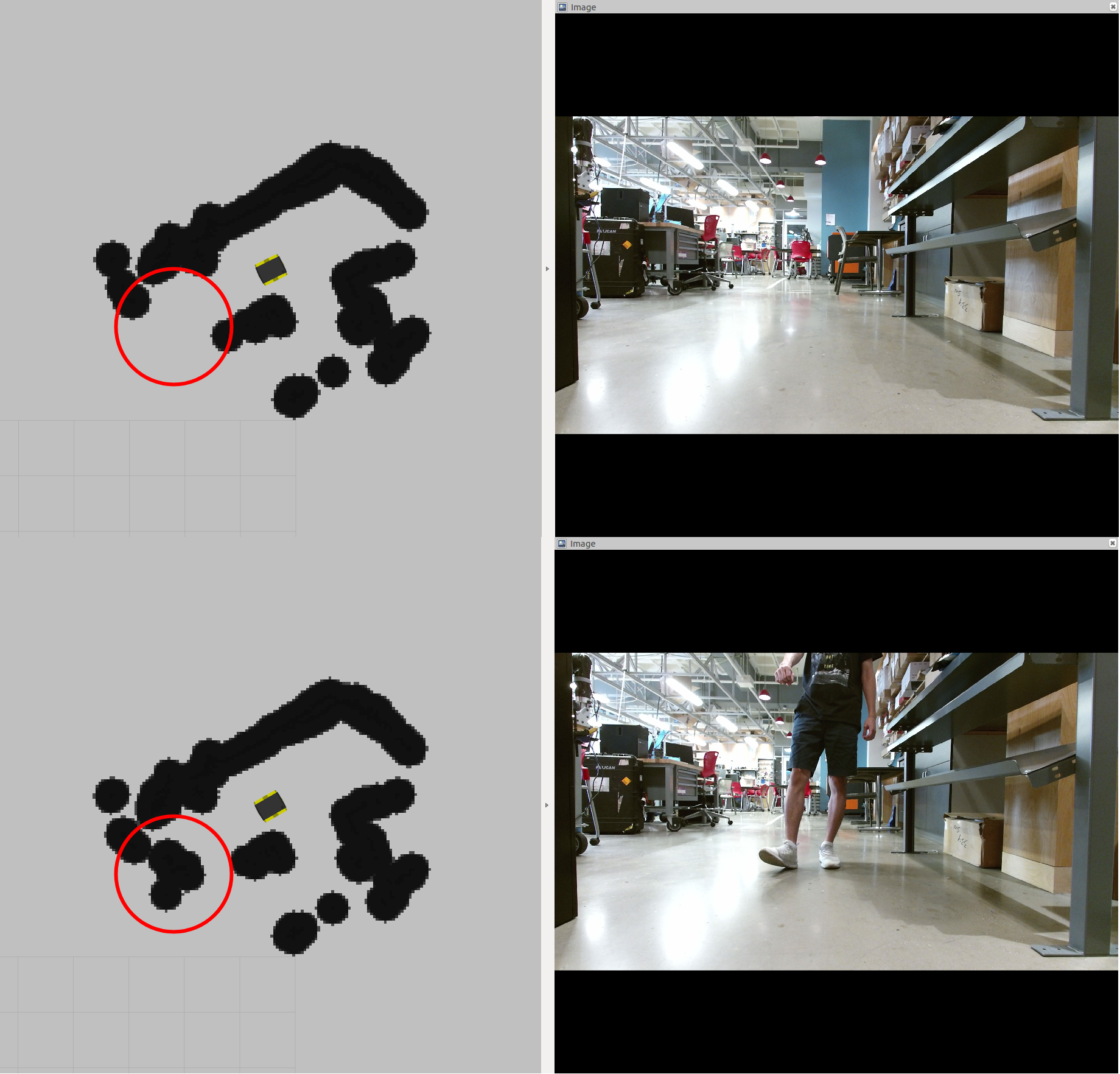}}
    \caption{Graphical User Interface (GUI) with and without a person (top \& bottom respectively) indicating how dynamic obstacles appear to the operator.}
    \label{fig:dynamic_map}
\end{figure}

Conventional approaches to shared control are usually done in the context of a known environment, where both the human operator and the autonomous agent are aware of the global map \cite{goil13}. Obstacles are static and there is normally a single path the operator takes. Our shared control architecture is tested in an environment unknown to both human and autonomous operators with dynamic obstacles, with Figure \ref{fig:dynamic_map} as an example. Furthermore, we exploit a robot with a defective IMU and weak wireless connection in a construction zone in an attempt to replicate a realistic situation.

In section \ref{sec:setup}, we validate the shared control architecture by simulating time delay and drift and compare the performance of the operator completing a navigation task with and without shared control. After validation, we observe the performance of the human-robot team when drift \& delay are no longer simulated. The operator and autonomous agent are only aware of their immediate surroundings in the form of a local costmap. In doing so, we provide a broader insight into dynamic human-robot team collaboration and how one could enhance the dynamics of such a team for better task completion.

% Add circle to show where map changes
% --------------------------------------------------------------------------------

% -------------------- %
% --- Architecture --- % 
% -------------------- %
\section{Proposed Shared Control Architecture}

\label{sec:implementation}
In our previous work \cite{aeroconf, dimitrov_roman}, we explored various schemes to consider and treat the human-robot team as a whole. This remains an open and exciting research area due to the challenges in modeling the uncertainties in human actions and hybrid nature of the system. In the following work, we consider the specific blended shared control approach presented in \cite{aeroconf}. 
We leverage the kinematic model of the system. Instead of formulating the user input in the control law for the system, we formulate the user input as a constraint on the kinematic model. Our formulation can handle a system that is redundant with respect to the task space dimensions, so the redundancy can be taken advantage of to implement a primary and secondary task. We integrate the human input to implement the shared control implementation in one of these tasks depending on the specific application. Our proposed architecture is summarized in Figure \ref{fig:sc_arch}.

\subsection{Formalism}
The primary task is a composition of both human and autonomous inputs described by 
\begin{align}
\dot{p}_p &= \alpha\dot{p}_u + (1-\alpha)\dot{p}_a \nonumber \\
		  &= \begin{bmatrix} 
		   \alpha\dot{x}_u + (1-\alpha)\dot{x}_a \\
		   \alpha\dot{y}_u + (1-\alpha)\dot{y}_a \\
		   \alpha\dot{\theta}_{u} + (1-\alpha)\dot{\theta}_{a}\end{bmatrix} 
\label{eqn:blended_primary}
\end{align}

\noindent where $\dot{p}_p$ designates the primary task velocity that the low-level robot controller will attempt to follow. $\dot{p}_u$ and $\dot{p}_a$ designate the user commanded input velocity and autonomous agent calculated velocity, respectively. The blending parameter $\alpha$ arbitrates the control of the system between user and autonomous input, and in our implementation is chosen as described in \cite{enes10}, shown in (\ref{eqn:enes_blend})
\begin{align}
\alpha &= \mbox{max}\left(0, 1-\frac{d}{d_0}\right) \cdot \mbox{max} \left( 0, 1-\left(\frac{\Delta}{\Delta_0}\right)^2\right)
\label{eqn:enes_blend}
\end{align}

\noindent where $d$ is the distance to the goal and $\Delta$ is the difference between the operator and autonomous agent's command, $\theta_u$ and $\theta_a$ respectively. $d_0$ and $\Delta_0$ are the maximum distances and commands, respectively, that the operator can deviate from.
Due to the motion of differential drive robots being constrained to the floor plane, the inputs and outputs can be represented by $\dot{x}$, $\dot{y}$, $\dot{\theta}$ corresponding to the 3 degrees of freedom on the plane.

A public release of the source code that implements this blended shared control framework can be found at https://github.com/piraka9011/jackal\_bsc. It can be used on most differential drive robots that use ROS.
% --------------------------------------------------------------------------------

% ------------------- %
% ----- Results ----- % 
% ------------------- %
\section{Experimental Setup}
\label{sec:setup}

In order to validate the blended shared control framework, we conducted two different studies. The first study, with 12 volunteers (ages from 18-24), asked users to teleoperate a Turtlebot around an obstacle and traverse a doorway. The task was considered complete if the Turtlebot passed a specific marker. Each volunteer performed 8 runs of the task: a manual run with no shared controller, a run with the shared controller, and 6 runs with the shared controller and either time delay or drift. The time delay here is the latency between when the user sends a teleoperation command and when the shared controller actually receives the command. Drift is simulated as a displacement in position between the odometry and map frames perceived by the robot, causing a difference in the calculated rotational velocity commanded to the robot. The order of the runs were randomized and the time delay and drift parameters are summarized in Table \ref{tab:door_experiment}. 

\begin{table}[ht]
    \centering
    \caption{Doorway Traversal Experiment Setup}
    \label{tab:door_experiment}
    \begin{tabular}{|ll|}
    \hline
    \multicolumn{2}{|c|}{\bf{Parameters}}      \\
    \hline
    Time Delay:  & \{0.5, 1.0, 2.0\} sec     \\
    Drift:       & \{0.1, 0.3, 0.5\} rad/sec \\
    \hline
    \multicolumn{2}{|c|}{\bf{Performance Metrics}} \\
    \hline
    Robot Odometry       & (meters)   \\
    Time to Completion   & (seconds)  \\
    \hline
    \end{tabular}
\end{table}

\noindent The environment was known to both the autonomous agent and operator. We use the distance the robot travels and the time for the operator to complete the task as performance metrics, shown in \ref{tab:door_experiment}. 

\begin{figure}[ht]
    \centering
    \includegraphics[width=\columnwidth]{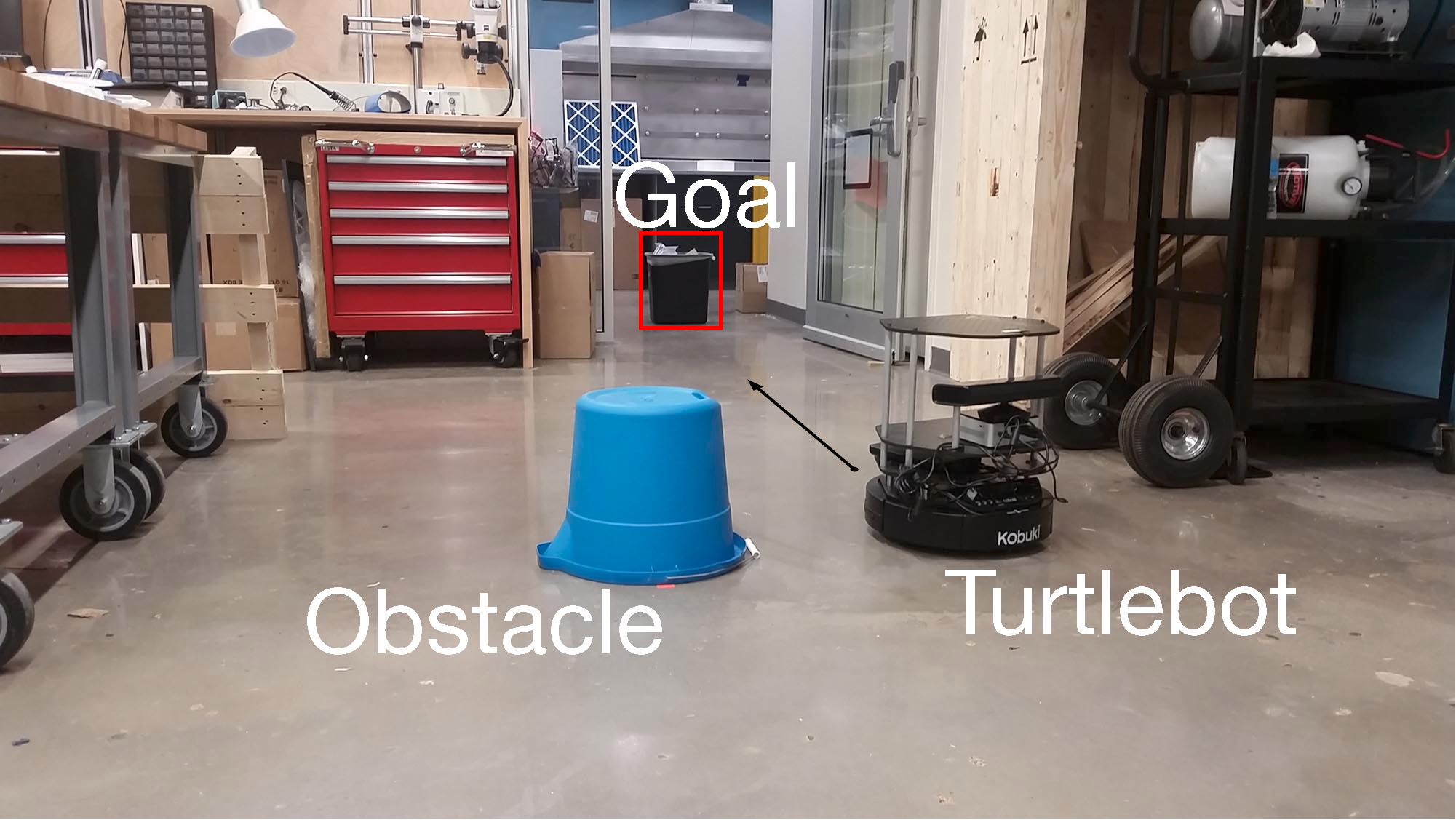}
    \caption{The blended shared control experiment for doorway traversal shown near the obstacle.}
    \label{fig:door_experiment}
\end{figure}

Figure \ref{fig:door_experiment} shows the setup for the doorway traversal experiment with a Turtlebot. The goal of the first study was to quantify the degradation of user performance as the simulated time delay and drift are varied and validate our shared control architecture. The time to complete the navigation task and distance traveled are recorded for each run, and used as metrics for comparing the performance of each run. 
The robot starts away from the door and needs to travel a short distance before turning left. Once the turn is complete, the robot needs to travel around the blue bucket shown in Figure \ref{fig:door_experiment}, and then through the door. The run ends when the robot reaches the location marked by a bin directly past the door. 

\begin{figure}[ht]
    \centering
    \includegraphics[width=\columnwidth]{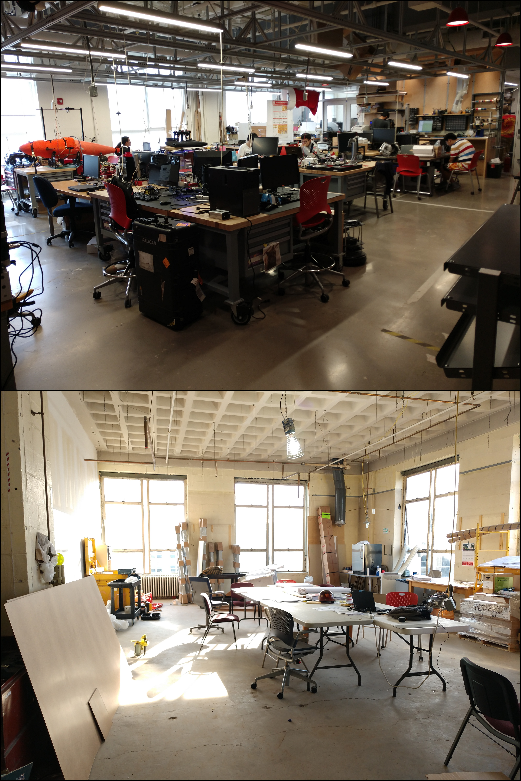}
    \caption{The lab (top) and construction (bottom) environments where users teleoperated the robot.}
    \label{fig:lab_const}
\end{figure}

The second study, with 14 volunteers (ages from 18-24), asked users to complete a complex navigation task in two different environments: a construction area with obstacles and a laboratory environment with tight corridors (Figure \ref{fig:lab_const}). The users had to drive the robot from a predefined starting position to a predefined goal position. The path to the goal is not clear and can be achieved via multiple routes, however, a marker similar to that used in the initial study is used to indicate the goal to the operator. 7 tests were completed in the construction area and 7 in the lab environment. Obstacle locations changed across tests and walking people were present in the environments. Volunteers were given a likert-scale questionnaire after the experiment regarding their performance and ability to control the robot. This study did not have simulated time delay and drift. Instead, we used a Clearpath Jackal with a defective IMU and weak wireless connection that caused a delay in user input and visual feedback. A Kinect V2 was attached to the Jackal and was used as the source of visual feedback.

The autonomous agent is implemented as a Robot Operating System (ROS) \cite{ros_paper} navigation stack that uses the dynamic window approach for local navigation \cite{ros_dwa_paper}. The autonomous agent is only given a goal with respect to the map frame and navigates using the robot's internal odometry. The autonomous agent does not have a global map of the environment, generating a navigation plan with respect to it's surrounding obstacles only.

The time to complete the navigation task was recorded for the individuals in all scenarios. The volunteers could see the robot but were asked to use the GUI similar to Figure \ref{fig:dynamic_map}. Due to the noisy odometry, we had no accurate means of recording the distance traveled. Thus, we used only time to completion as a performance metric along with the likert-scale questionnaire. The speeds in both studies were set to a constant 1.0 m/s across all tests and the shared control parameters were chosen to be $\Delta_0 = 3.0\: m/s$ and $d_0 = 15.0\: m$. Although we could not control for delay and odometry noise in the second study, the delay had a range of 0.5s-1.5s. This was measured using the difference in wall clock time and when messages were received by the Clearpath Jackal in ROS time.

% --------------------------------------------------------------------------------

% ------------------- %
% ----- Results ----- % 
% ------------------- %
\section{Validation Study Results}
\label{sec:validation_results}

\begin{figure}[h!]
    \centering
    \includegraphics[width=\columnwidth]{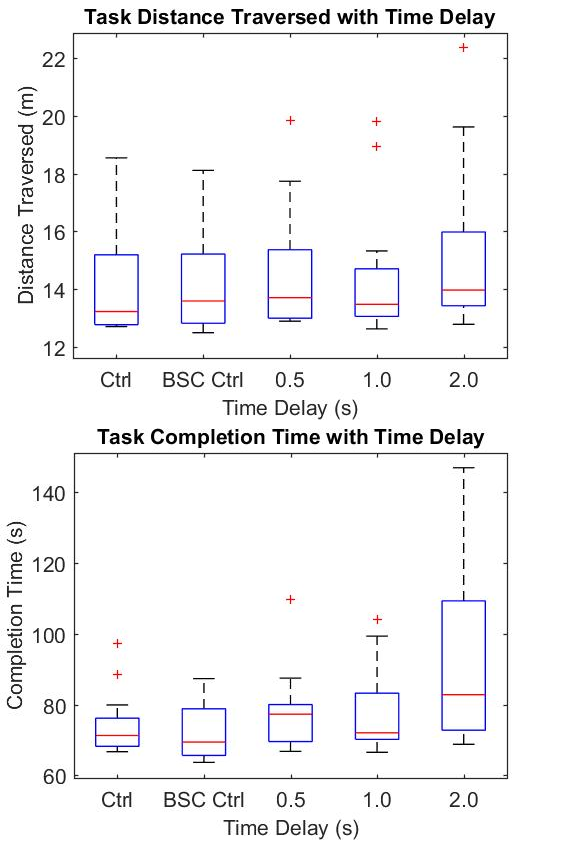}
    \caption{Boxplots of distances traversed (top) and completion times (bottom) for experimental trials with drift.}
    \label{fig:boxplots_delay}
\end{figure}

\begin{figure}[ht]
    \centering
    \includegraphics[width=\columnwidth]{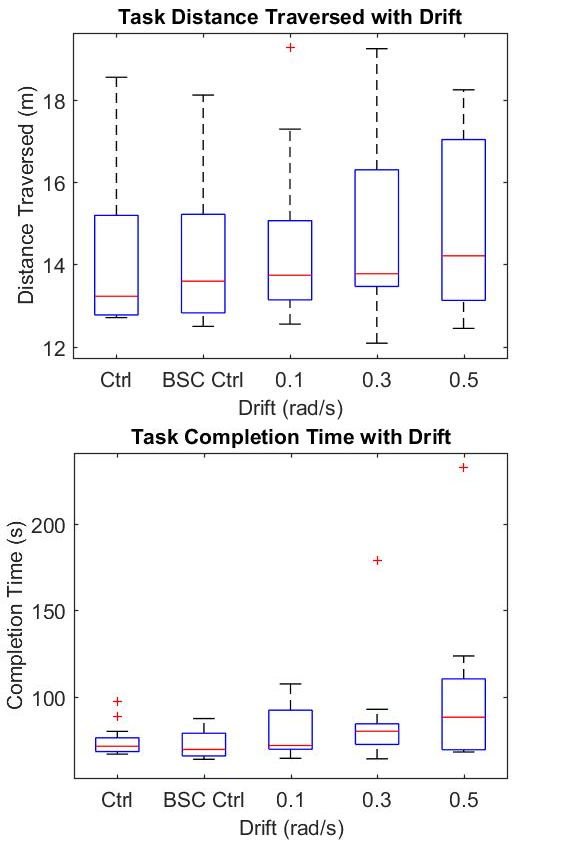}
    \caption{Boxplots of distances traversed (top) and completion times (bottom) for experimental trials with delay.}
    \label{fig:boxplots_drift}
\end{figure}

Here we present the results of our studies and our observations. For each scenario, we determine if there is a statistical difference in performance across trials by running a Wilcoxon Signed-Rank test. All hypotheses are tested at the 5\% significance level ($p<0.05$). Figures \ref{fig:boxplots_drift} and \ref{fig:boxplots_delay} summarize the results for the first study. To evaluate the statistical significance of operator performance, we formally define our hypotheses as follows. 

\subsection{Hypotheses}
\begin{nullhyp}
The times to complete the doorway traversal task under the blended shared control scenario and the time delay scenarios come from the same distribution.
\end{nullhyp}

The null hypothesis is rejected for all three time delay values (0.5, 1.0, and 2.0 sec) with respective $p$-value's of 0.0269, 0.0269, and 4.88e-04. This result indicates that all time delay scenario data does not come from the same distribution as the blended shared control scenario, meaning the result is in fact statistically significant. Users took longer to complete the doorway traversal task with time delay disturbances than in the blended shared control scenario.

\begin{nullhyp}
The distances to complete the doorway traversal task under the blended shared control scenario and the time delay scenarios come from the same distribution.
\end{nullhyp}

The null hypothesis is accepted for the 0.5 and 1.0 second delay parameters, and rejected for 2.0 second delay with respective $p$-values of 0.33, 0.56, 0.0093. This result indicates that the distance metric is not affected as much by the time metric in the time delay scenarios. Not until the delay becomes quite large, does the change in distance traversed become statistically significant.

\begin{nullhyp}
The distances to complete the doorway traversal task under the blended shared control scenario and the drift scenarios come from the same distribution.
\end{nullhyp}

The null hypothesis is accepted for all drift scenarios (0.1, 0.3, 0.5 rad/sec) with $p$-value's of 0.62, 0.17, 0.10 respectively. This result indicates that the data is inconclusive and would require more study to reject the null hypotheses. The constraints posed by the specific path needed to complete the task likely contribute to this result.

\begin{nullhyp}
The times to complete the doorway traversal task under the blended shared control scenario and the drift scenarios come from the same distribution.
\end{nullhyp}

It is interesting to note the null hypothesis for time is accepted for the 0.1 rad/sec and rejected for 0.3 and 0.5 rad/sec scenarios with $p$-values 0.51, 0.034, and 4.88e-04. This result indicates that despite the insignificant change in distance to complete the task, the task did take longer to complete with the drift. Both the obstacle and doorway limited the paths the operators could take, and thus may have limited the variability in the distances.

\subsection{Discussion}
The results indicate that users took longer to complete the task with time-delay, as expected. However, it is interesting to note that qualitative testing with the time-delay scenario and no blended shared controller led to a robot that was almost impossible to accurately control through teleoperation. The blended shared controller enables the users to traverse the doorway, albeit at a slower pace, but nevertheless successfully. In the context of a search \& rescue task, this latency would be detrimental to both task completion and the integrity of the robot, where multiple collisions could occur. Blending the autonomous agent's input with the operator's makes task completion feasible and the robot's behavior more reliable.

% --------------------------------------------------

% ------------------- %
% ----- Results ----- % 
% ------------------- %
\section{Dynamic Environment Results}
\label{sec:dynamic_results}
Following a similar analysis to the validation study, we perform a Wilcoxon signed rank test on the times taken to complete the task. Since we cannot control for delay and drift, we formulate a hypothesis based on the times taken to complete the task manually and with shared control only. 

\subsection{Hypotheses}

\begin{figure}[ht]
    \centering
    \includegraphics[width=\columnwidth]{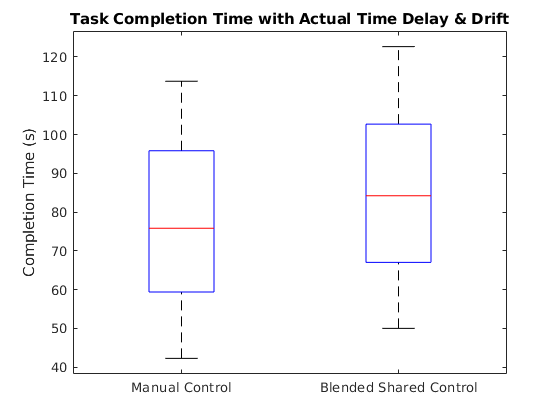}
    \caption{Boxplot of time to task completion when both input delay and odometry drift are present in a dynamic environment for manual and blended shared control.}
    \label{fig:dyn_boxplot}
\end{figure}

\begin{nullhyp}
The times to complete the navigation task under the blended shared control and the manual control scenarios in all environments come from the same distribution.
\end{nullhyp}

The null hypothesis is accepted for both construction and lab environment navigation tasks with $p$-value's of 0.8125 and 0.0938 and respectively. Thus, we can conclude that navigation with shared control did not hamper the performance of the operator under manual control. A boxplot of the results are shown in Figure \ref{fig:dyn_boxplot}.

\subsection{Post-Experiment Survey}
\begin{figure}[ht]
    \centering
    \includegraphics[width=.9\columnwidth]{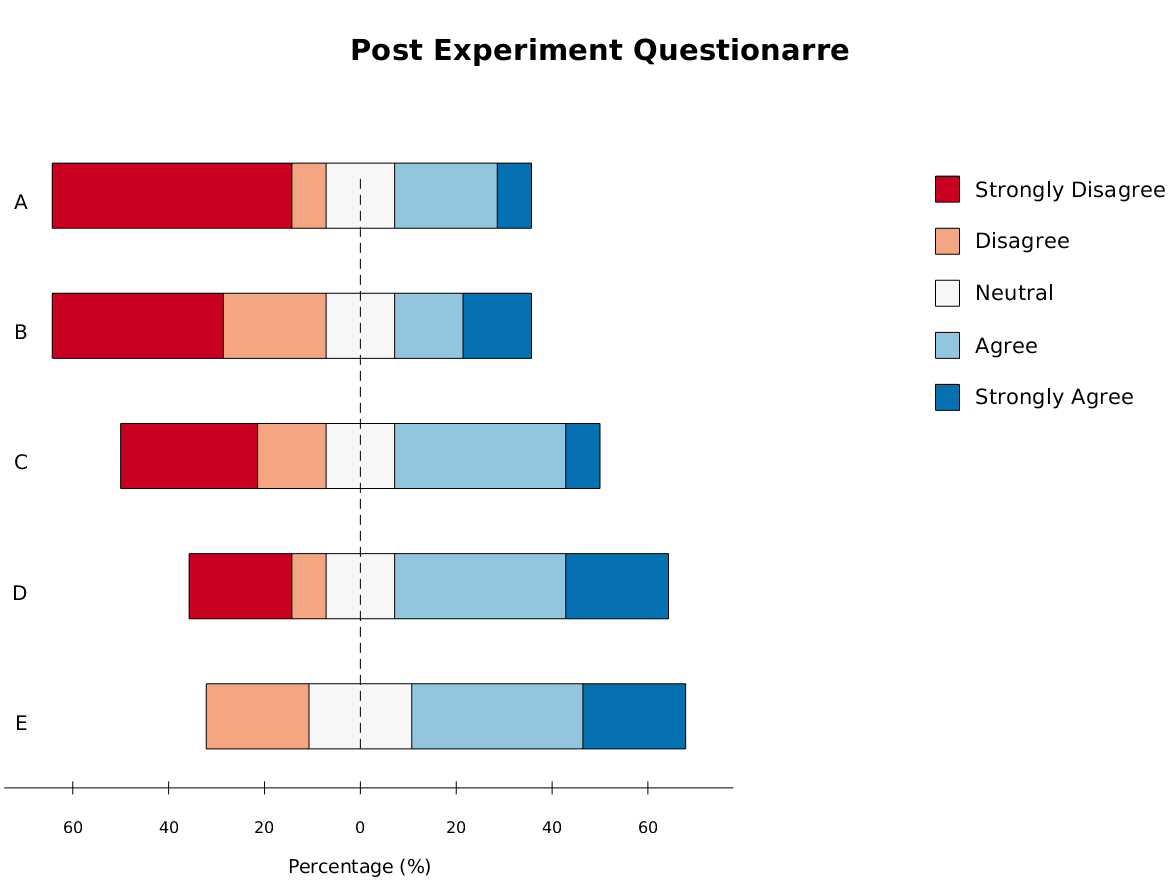}
    \caption{Likert scale for post experiment questionnaire.}
    \label{fig:likert}
\end{figure}

\begin{table}[ht]
    \caption{Post experiment survey questions}
    \label{tab:survey_questions}
    \begin{center}
        \begin{tabularx}{\columnwidth}{ | l | X | r | }
        \hline
        \textbf{Label} & \textbf{Question} & \textbf{Mean}\\
        \hline
        A & I experienced a delay in input/steering.     & $4.07 \pm 2.92$ \\
        \hline
        B & I was frustrated attempting to complete the task.    & $4.43 \pm 3.00$ \\
        \hline
        C & I was unable to perform to the best of my abilities. & $5.07 \pm 2.87$ \\
        \hline
        D &  I experienced a delay in visual feedback.    & $5.93 \pm 3.15 $ \\
        \hline
        E & I felt that I had control over the robot.    & $6.64 \pm 2.24$ \\
        \hline
        \end{tabularx}
    \end{center}
\end{table}

A post experiment survey was conducted to understand how operators felt about their performance and whether they were comfortable with the assistance of the shared controller. It also gave us insight as to whether users experienced delay during their runs or not. Results are summarized in Table \ref{tab:survey_questions} and Figure \ref{fig:likert}. Scores of 10 and 1 indicate that the responder strongly agrees or disagrees, respectively.

\subsection{Discussion}
All users successfully completed the task with no collisions. Most users took similar routes that corresponded with the navigation stack's route. This is most likely caused by the \textit{resistance} user's felt from the autonomous agent when attempting to take a different route. When users did not take the suggested route, the navigation stack would recalculate the path and the autonomous agent would essentially \textit{comply} with the user's decision. This behavior was interesting to observe when the autonomous agent suggested a route that the operator knew was erroneous. Clearly, there was a human-robot team dynamic occurring that warrants further investigation towards the choice of the shared control parameter.

One method that could limit the resistance between the human operator and autonomous agent is redefining the blending parameter $\alpha$ to consider only the distance to the goal and the euclidean distance to the nearest obstacle. This would mimic the behavior of a potential field based controller, potentially enhancing both the performance of the human-robot team and the feeling of the operator in control.

The survey suggests that although operators felt they were in control of the robot, they were frustrated in attempting to reach the goal. This is most likely caused by our choice of $\alpha$ which relinquishes control to the autonomous agent as the goal is approached. The autonomous agent would resist the operators' attempt to correct their heading as they approached the goal pose, with a limited effect on the behavior of the robot. Relaxing this constraint may allow the operator to more freely reach the desired goal pose, but it would minimize the autonomous agent's intervention should the operator stray away from the goal.

We cannot draw concrete conclusions from the survey regarding whether users felt they were able to perform to the best of their abilities. However, the wide range of responses indicates that some users saw the shared controller as an inhibitor of performance, while others saw it as an aid. 

The most important observation is that despite there being an input delay due to the weak wireless connection, the survey suggests that operators experiencing no such delay. The blended shared controller  would correct the robot's heading even when the operator's input lagged, enhancing teleoperation of the robot. This is very important in the context of situational awareness as it suggests that operators remained aware of their environment and could navigate the robot with ease. Complementing this with the fact that operators felt they were in control is a step forward in designing architectures for amicable human-robot teams.
% --------------------------------------------------------------------------------

% ------------------- %
% --- Future Work --- % 
% ------------------- %
\section{Future Work}
\label{sec:futurework}
Future work will focus on improving the blended shared control algorithm through two key areas: the selection of $\alpha$, either by changing the model or using reinforcement learning, and utilizing higher complexity systems. While the experiments show that our selection of $\alpha$ results in good performance, there may be better ways to define the parameter and its associated constants. Considerations should include how operators like the \emph{feel} of the change in control authority over the system and the context in which it is used. The blending parameter may for example prioritize multiple goal poses if the task is trajectory based. The shared control architecture can also be extended through reinforcement learning, where an agent could learn the shared control parameter using a reward function based on the context it is used in.

Another area to explore is how the blended shared control architecture changes human-robot team dynamics with different environments and tasks (ex. exploration vs. goal locations). Further research also needs to be done on the feasibility of such an architecture with manipulation tasks in high degree-of-freedom systems, such as humanoid robots. Specifically, whether performance is enhanced when blended shared control is applied in joint space or end-effector space.
% --------------------------------------------------------------------------------

\section{Conclusion}
\label{sec:conclusion}
We presented an analysis on the performance of a blended human-robot shared control architecture with operator input latency and odometry drift. The architecture is initially validated in a study with simulated latency and drift with human operators. It is then tested in a second study in multiple dynamic environments with a robot that suffers from actual latency and drift. Quantitative results for both studies show improved operator performance in terms of time to task completion with the shared controller as opposed to no shared controller when latency and drift are present. Future investigations into learning the blending parameter for architecture abstraction and potentially enhanced performance of a human-robot team is suggested. 

\bibliographystyle{IEEEtran}
\bibliography{velin,mendeley_v2}

\begin{thebibliography}{10}
\providecommand{\url}[1]{#1}
\csname url@rmstyle\endcsname
\providecommand{\newblock}{\relax}
\providecommand{\bibinfo}[2]{#2}
\providecommand\BIBentrySTDinterwordspacing{\spaceskip=0pt\relax}
\providecommand\BIBentryALTinterwordstretchfactor{4}
\providecommand\BIBentryALTinterwordspacing{\spaceskip=\fontdimen2\font plus
\BIBentryALTinterwordstretchfactor\fontdimen3\font minus
  \fontdimen4\font\relax}
\providecommand\BIBforeignlanguage[2]{{%
\expandafter\ifx\csname l@#1\endcsname\relax
\typeout{** WARNING: IEEEtran.bst: No hyphenation pattern has been}%
\typeout{** loaded for the language `#1'. Using the pattern for}%
\typeout{** the default language instead.}%
\else
\language=\csname l@#1\endcsname
\fi
#2}}

\bibitem{sheridan:92}
T.~B. Sheridan, \emph{Telerobotics, Automation, and Human Supervisory
  Control}.\hskip 1em plus 0.5em minus 0.4em\relax Cambridge, MA, USA: MIT
  Press, 1992.

\bibitem{hayati:89}
S.~Hayati and S.~Venkataraman, ``{Design and Implementation of a Robot Control
  System with Traded and Shared Control Capability},'' in \emph{Robotics and
  Automation, 1989. Proceedings., 1989 IEEE International Conference on}, May
  1989, pp. 1310 --1315 vol.3.

\bibitem{anderson:96}
R.~Anderson, ``{Autonomous, Teleoperated, and Shared control of Robot
  Systems},'' in \emph{Robotics and Automation, 1996. Proceedings., 1996 IEEE
  International Conference on}, vol.~3, April 1996, pp. 2025 --2032 vol.3.

\bibitem{lacey:00}
\BIBentryALTinterwordspacing
G.~Lacey and S.~MacNamara, ``{Context-Aware Shared Control of a Robot Mobility
  Aid for the Elderly Blind},'' \emph{The International Journal of Robotics
  Research}, vol.~19, no.~11, pp. 1054--1065, 2000. [Online]. Available:
  \url{http://ijr.sagepub.com/content/19/11/1054.abstract}
\BIBentrySTDinterwordspacing

\bibitem{crandall:02}
J.~Crandall and M.~Goodrich, ``{Characterizing Efficiency of Human Robot
  Interaction: A Case Study of Shared-Control Teleoperation},'' in
  \emph{Intelligent Robots and Systems, 2002. IEEE/RSJ International Conference
  on}, vol.~2, 2002, pp. 1290 -- 1295 vol.2.

\bibitem{wasson:03}
G.~Wasson, P.~Sheth, M.~Alwan, K.~Granata, A.~Ledoux, and C.~Huang, ``{User
  Intent in a Shared Control Framework for Pedestrian Mobility Aids},'' in
  \emph{Intelligent Robots and Systems, 2003. (IROS 2003). Proceedings. 2003
  IEEE/RSJ International Conference on}, vol.~3, October 2003, pp. 2962 -- 2967
  vol.3.

\bibitem{simpson:97}
R.~Simpson and S.~Levine, ``{Adaptive Shared Control of a Smart Wheelchair
  Operated by Voice Control},'' in \emph{Intelligent Robots and Systems, 1997.
  IROS '97., Proceedings of the 1997 IEEE/RSJ International Conference on},
  vol.~2, September 1997, pp. 622 --626 vol.2.

\bibitem{gopinath17}
D.~Gopinath, S.~Jain, and B.~D. Argall, ``Human-in-the-loop optimization of
  shared autonomy in assistive robotics,'' \emph{IEEE Robotics and Automation
  Letters}, vol.~2, no.~1, pp. 247--254, Jan 2017.

\bibitem{Broad2017LearningDynamics}
A.~Broad, T.~Murphey, and B.~Argall, ``{Learning Models for Shared Control of
  Human-Machine Systems with Unknown Dynamics},'' in \emph{Robotics: Science
  and Systems}, 2017.

\bibitem{Yu2003AnElderly}
H.~Yu, M.~Spenko, and S.~Dubowsky, ``{An Adaptive Shared Control System for an
  Intelligent Mobility Aid for the Elderly},'' \emph{Autonomous Robots},
  vol.~15, pp. 53--66, 2003.

\bibitem{Yanco2004quotWherePlatform}
H.~A. Yanco and J.~Drury, ``{{\&}quot;Where Am I?{\&}quot; Acquiring Situation
  Awareness Using a Remote Robot Platform},'' in \emph{IEEE International
  Conference on Systems, Man and Cybernetics}.\hskip 1em plus 0.5em minus
  0.4em\relax The Hague: IEEE, 2004.

\bibitem{Kartoun2010AAlgorithm}
U.~Kartoun, H.~Stern, and Y.~Edan, ``{A human-robot collaborative reinforcement
  learning algorithm},'' \emph{Journal of Intelligent and Robotic Systems:
  Theory and Applications}, 2010.

\bibitem{Doroodgar2010TheRobots}
B.~Doroodgar, M.~Ficocelli, B.~Mobedi, and G.~Nejat, ``{The search for
  survivors: Cooperative human-robot interaction in search and rescue
  environments using semi-autonomous robots},'' in \emph{Proceedings - IEEE
  International Conference on Robotics and Automation}, 2010.

\bibitem{Doroodgar2010AEnvironments}
B.~Doroodgar and G.~Nejat, ``{A Hierarchical Reinforcement Learning Based
  Control Architecture for Semi-Autonomous Rescue Robots in Cluttered
  Environments},'' in \emph{International Conference on Automation Science and
  Engineering}.\hskip 1em plus 0.5em minus 0.4em\relax Toronto: IEEE, 2010.

\bibitem{Reddy2018SharedLearning}
S.~Reddy, A.~D. Dragan, and S.~Levine, ``{Shared Autonomy via Deep
  Reinforcement Learning},'' in \emph{Robotics: Science and Systems}, 2018.

\bibitem{goil13}
A.~Goil, M.~Derry, and B.~D. Argall, ``Using machine learning to blend human
  and robot controls for assisted wheelchair navigation,'' in \emph{2013 IEEE
  13th International Conference on Rehabilitation Robotics (ICORR)}, June 2013,
  pp. 1--6.

\bibitem{aeroconf}
V.~Dimitrov and T.~Padir, ``A comparative study of teleoperated and autonomous
  task completion for sample return rover missions,'' in \emph{Aerospace
  Conference, 2014 IEEE}, March 2014, pp. 1--6.

\bibitem{dimitrov_roman}
V.~Dimitrov and T.~Padır, ``A shared control architecture for
  human-in-the-loop robotics applications,'' in \emph{The 23rd IEEE
  International Symposium on Robot and Human Interactive Communication}, Aug
  2014, pp. 1089--1094.

\bibitem{enes10}
A.~Enes and W.~Book, ``Blended shared control of zermelo's navigation
  problem,'' in \emph{American Control Conference (ACC), 2010}, June 2010, pp.
  4307--4312.

\bibitem{ros_paper}
M.~Quigley, K.~Conley, B.~P. Gerkey, J.~Faust, T.~Foote, J.~Leibs, R.~Wheeler,
  and A.~Y. Ng, ``Ros: an open-source robot operating system,'' in \emph{ICRA
  Workshop on Open Source Software}, 2009.

\bibitem{ros_dwa_paper}
D.~Fox, W.~Burgard, and S.~Thrun, ``The dynamic window approach to collision
  avoidance,'' \emph{Robotics Automation Magazine, IEEE}, vol.~4, no.~1, pp.
  23--33, Mar 1997.

\end{thebibliography}

\end{document}